\documentclass{article}

\usepackage{arxiv}

\usepackage[utf8]{inputenc} % allow utf-8 input
\usepackage[T1]{fontenc}    % use 8-bit T1 fonts
\usepackage{hyperref}       % hyperlinks
\usepackage{url}            % simple URL typesetting
\usepackage{booktabs}       % professional-quality tables
\usepackage{amsfonts}       % blackboard math symbols
\usepackage{nicefrac}       % compact symbols for 1/2, etc.
\usepackage{microtype}      % microtypography
\usepackage{lipsum}
\usepackage{amsmath,amssymb,amsfonts, algorithm2e}
\usepackage{pifont}% http://ctan.org/pkg/pifont
\newcommand{\cmark}{\ding{51}}%
\newcommand{\xmark}{\ding{55}}%\textbf{}
\usepackage{algorithmic}
\usepackage{graphicx}
\usepackage{textcomp}
\usepackage{xcolor}

\usepackage{multirow}

\title{PointFlowHop: Green and Interpretable Scene Flow Estimation from Consecutive Point Clouds}

\author{
Pranav Kadam \\
Media Communications Lab \\
University of Southern California\\
Los Angeles, CA, USA \\
\texttt{pranavka@usc.edu} \\
\And
Jiahao Gu \\
Media Communications Lab\\
University of Southern California\\
Los Angeles, CA, USA \\
\texttt{jiahaogu@usc.edu} \\
  
\And
Shan Liu \\
Tencent Media Lab\\
Tencent America\\
Palo Alto, CA, USA \\
\texttt{shanl@tencent.com} \\
  
\And
C.-C. Jay Kuo \\
Media Communications Lab\\
University of Southern California\\
Los Angeles, CA, USA \\
\texttt{cckuo@sipi.usc.edu} \\
}

\begin{document}
\maketitle

\begin{abstract}
An efficient 3D scene flow estimation method called PointFlowHop is
proposed in this work. PointFlowHop takes two consecutive point clouds
and determines the 3D flow vectors for every point in the first point
cloud. PointFlowHop decomposes the scene flow estimation task into a set
of subtasks, including ego-motion compensation, object association and
object-wise motion estimation. It follows the green learning (GL)
pipeline and adopts the feedforward data processing path.  As a result,
its underlying mechanism is more transparent than deep-learning (DL)
solutions based on end-to-end optimization of network parameters.  We
conduct experiments on the stereoKITTI and the Argoverse LiDAR point
cloud datasets and demonstrate that PointFlowHop outperforms
deep-learning methods with a small model size and less training time.
Furthermore, we compare the Floating Point Operations (FLOPs) required
by PointFlowHop and other learning-based methods in inference, and show
its big savings in computational complexity. 
\end{abstract}


\section{Introduction}\label{sec:introduction}

Dynamic 3D scene understanding based on captured 3D point cloud data is
a critical enabling technology in the 3D vision systems. 3D scene flow
aims at finding the point-wise 3D displacement between consecutive point
cloud scans.  With the increase in the availability of point cloud data,
especially those acquired via the LiDAR scanner, 3D scene flow
estimation directly from point clouds is an active research topic
nowadays. 3D scene flow estimation finds rich applications in 3D
perception tasks such as semantic segmentation, action recognition, and
inter-prediction in compressing sequences of LiDAR scans. 

Today's solutions to 3D scene flow estimation mostly rely on supervised
or self-supervised deep neural networks (DNNs) that learn to predict the
point-wise motion field from a pair of input point clouds via end-to-end
optimization. One of the important components of these methods is to
learn flow embedding by analyzing spatio-temporal correlations among
regions of the two point clouds.  After the successful demonstration of
such an approach in FlowNet3D \cite{liu2019flownet3d}, there has been an
increased number of papers on this topic by exploiting and combining
other ideas such as point convolutions and attention mechanism. 

These DNN-based methods work well in an environment that meets the local
scene rigidity assumption. They usually outperform classical
point-correspondence-based methods.  On the other hand, they have a
large number of parameters and rely on large training datasets.  For the
3D scene flow estimation problem, it is non-trivial to obtain dense
point-level flow annotations. Thus, it is challenging to adopt the
heavily supervised learning paradigm with the real world data.  Instead,
methods are typically trained on synthetic datasets with ground truth
flow information first. They are later fine-tuned for real world
datasets. This makes the training process very complicated. 

In this paper, we develop a green and interpretable 3D scene flow
estimation method for the autonomous driving scenario and name it
``PointFlowHop''. We decompose our solution into vehicle ego-motion and
object motion modules. Scene points are classified as static and moving.
Moving points are grouped into moving objects and a rigid flow model is
established for each object. Furthermore, the flow in local regions is
refined assuming local scene rigidity.  PointFlowHop method adopts the
green learning (GL) paradigm \cite{kuo2022green}.  It is built upon
related recent work, GreenPCO \cite{kadam2022greenpco}, and preceding
foundation works such as R-PointHop \cite{kadam2022r}, PointHop
\cite{zhang2020pointhop}, and PointHop++ \cite{zhang2020pointhop++}. 

The task-agnostic nature of the feature learning process in prior art
enables scene flow estimation through seamless modification and
extension. Furthermore, a large number of operations in PointFlowHop are
not performed during training. The ego-motion and object-level motion is
optimized in inference only. Similarly, the moving points are grouped
into objects only during inference. This makes the training process much
faster and the model size very small.  The decomposition of 3D scene
flow into object-wise rigid motion and/or ego-motion components is not
entirely novel. However, our focus remains in developing a GL-based
solution with improved overall performance, including accuracy, model
sizes and computational complexity. 

The novelty of our work lies in two aspects. First, it expands the scope
of existing GL-based point cloud data processing techniques.  GL-based
point cloud processing has so far been developed for object-level
understanding \cite{kadam2023s3i, kadam2020unsupervised, kadam2022pcrp,
zhang2020unsupervised, zhang2020pointhop++, zhang2020pointhop} and
indoor scene understanding \cite{kadam2022r, zhang2022gsip}. This work
addresses the more challenging problem of outdoor scene understanding at
the point level. This work also expands the application scenario of
R-PointHop, where all points are transformed using one single rigid
transformation. For 3D scene flow estimation, each point has its own
unique flow vector. Furthermore, we show that a single model can learn
features for ego-motion estimation as well as object-motion estimation,
which are two different but related tasks. This allows model sharing and
opens doors to related tasks such as joint scene flow estimation and
semantic segmentation. Second, our work highlights the over-paramertized
nature of DL-based solutions which demand larger model sizes and higher
computational complexity in both training and testing. The overall
performance of PointFlowHop suggests a new point cloud processing
pipeline that is extremely lightweight and mathematically transparent. 

To summarize, there are three major contributions of this work. 
\begin{itemize}
\item We develop a lightweight 3D scene classifier that identifies
moving points and further clusters and associates them into moving
object pairs. 
\item We optimize the vehicle ego-motion and object-wise motion based 
on point features learned using a single task-agnostic feedforward
PointHop++ model. 
\item PointFlowHop outperforms representative benchmark methods in the
scene flow estimation task on two real world LiDAR datasets with fewer
model parameters and lower computational complexity measured by FLOPs
(floating-point operations). 
\end{itemize}

The rest of the paper is organized as follows.  Related work is reviewed
in Sec. \ref{sec:review}. The PointFlowHop method is proposed in Sec.
\ref{sec:method}. Experimental results are presented in Sec.
\ref{sec:experiments}.  Finally, concluding remarks and possible future
extensions are given in Sec. \ref{sec:conclusion}. 

\section{Related Work}\label{sec:review}

\subsection{Scene Flow Estimation}

Early work on 3D scene flow estimation uses 2D optical flow estimation
followed by triangulation such as that given in \cite{vedula1999three}.
The Iterative Closest Point (ICP) \cite{besl1992method} and the
non-rigid registration work, NICP \cite{amberg2007optimal}, can operate
on point clouds directly.  Series of image- and point-based seminal
methods for scene flow estimation relying on similar ideas were proposed
in the last two decades. The optical flow is combined with dense stereo
matching for flow estimation in \cite{huguet2007variational}. A
variational framework that predicts the scene flow and depth is proposed
in \cite{basha2013multi}.  A peicewise rigid scene flow estimation
method is investigated in \cite{vogel2013piecewise}.  Similarly, the
motion of rigidly moving 3D objects is examined in
\cite{menze2015object}.  Scene flow based on Lucas-Kanade tracking
\cite{lucas1981iterative} is studied in \cite{quiroga2012scene}. An
exhaustive survey on 2D optical flow and 3D scene flow estimation
methods has been done by Zhai et al.  \cite{zhai2021optical}. 

Deep-learning-based (DL-based) methods have been popular in the field of
computer vision in the last decade. For DL-based 3D scene flow
estimation, FlowNet3D \cite{liu2019flownet3d} adopts the feature
learning operations from PointNet++ \cite{qi2017pointnet++}. HPLFlowNet
\cite{gu2019hplflownet} uses bilateral convolution layers and projects
point clouds to an ordered permutohedral lattice.  PointPWC-Net
\cite{wu2020pointpwc} takes a self-supervised learning approach that
works in a coarse-to-fine manner.  FLOT \cite{puy2020flot} adopts a
correspondence-based approach based on optimal transport. HALFlow
\cite{wang2021hierarchical} uses a hierarchical network structure with
an attention mechanism. The Just-Go-With-the-Flow method
\cite{mittal2020just} uses self-supervised learning with the nearest
neighbor loss and the cycle consistency loss. 

DL-based methods that attempt to simplify the flow estimation problem
using ego-motion and/or object-level motion have also been investigated.
For example, Rigid3DSceneFlow \cite{gojcic2021weakly} reasons the scene
flow at the object level (rather than the point level).  Accordingly,
the flow of scene background is analyzed via ego-motion and that of a
foreground object is described by a rigid model. RigidFlow
\cite{li2022rigidflow} enforces the rigidity constraint in local regions
and performs rigid alignment in each region to produce rigid pseudo
flow.  SLIM \cite{baur2021slim} uses a self-supervised loss function to
separate moving and stationary points. 

\subsection{Green Point Cloud Learning}

Green Learning (GL) \cite{kuo2022green} has started to gain attention as
an alternative to Deep Learning (DL) in recent years. Typcially, GL
consists of three modules: 1) unsupervised representation learning, 2)
supervised feature learnaing, and 3) supervised decision learning.  The
unsupervised representation learning in the first module is rooted in
the derivation of data-driven transforms such as the Saak
\cite{kuo2018data} and the Saab \cite{kuo2019interpretable} transforms.
Both the training and inference processes in GL adopt a feedforward data
processing path without backpropagation. The optimization is
statistics-based, and it is carried out at each module independently.
The learning process is lightweight, making it data and computation
resource friendly.  GL-based methods have been developed for a wide
variety of image processing and computer vision tasks
\cite{kuo2022green}. 

Green Point Cloud learning \cite{liu20213d} was first introduced in
PointHop \cite{zhang2020pointhop}. The unsupervised representation
learning process involves constructing a local point descriptor via
octant space partitioning followed by dimensionality reduction via the
Saab transform. These operations together are called one PointHop unit.
It is the fundamental building block in series of follow-up works along
with other task-specific modules. PointHop++ \cite{zhang2020pointhop++}
replaces the Saab transform with its efficient counterpart called the
Channel-wise Saab transform \cite{chen2020pixelhop++}. PointHop and
PointHop++ adopt a multi-hop learning system for point cloud
classification, whereby the learned point representations are aggregated
into a global feature vector and fed to a classifier. The multi-hop
learning architecture is analogous to the hierarchical deep learning
architecture. The multi-hop architecture helps capture the information
from short-, mid-, and long-range point cloud neighborhoods. 

More recently, R-PointHop \cite{kadam2022r}, GSIP \cite{zhang2022gsip}
and GreenPCO \cite{kadam2022greenpco} demonstrate green learning
capabilities on more challenging large-scale point clouds for indoor
scene registration, indoor segmentation, and odometry tasks,
respectively. R-PointHop finds corresponding points between the source
and target point clouds using the learned representations and then
estimates the 3D rotation and translation to align the source with the
target. In GreenPCO, a similar process is adopted to incrementally
estimate the object's trajectory. Additional ideas presented in GreenPCO
include a geometry-aware point cloud sampling scheme that is suitable
for LiDAR data. Other noteworthy green point cloud learning works
include SPA \cite{kadam2020unsupervised}, UFF
\cite{zhang2020unsupervised}, PCRP \cite{kadam2022pcrp}, and
S3I-PointHop \cite{kadam2023s3i}. 

%%%%%%%%%%%%%%%%%%%%%%%%%%%%%%%%%%%%%%%%%%%%%%%%%%%%%%%%%%%%%%%%%%%%%
\begin{figure*}[!t]
\centering
\includegraphics[scale=0.45]{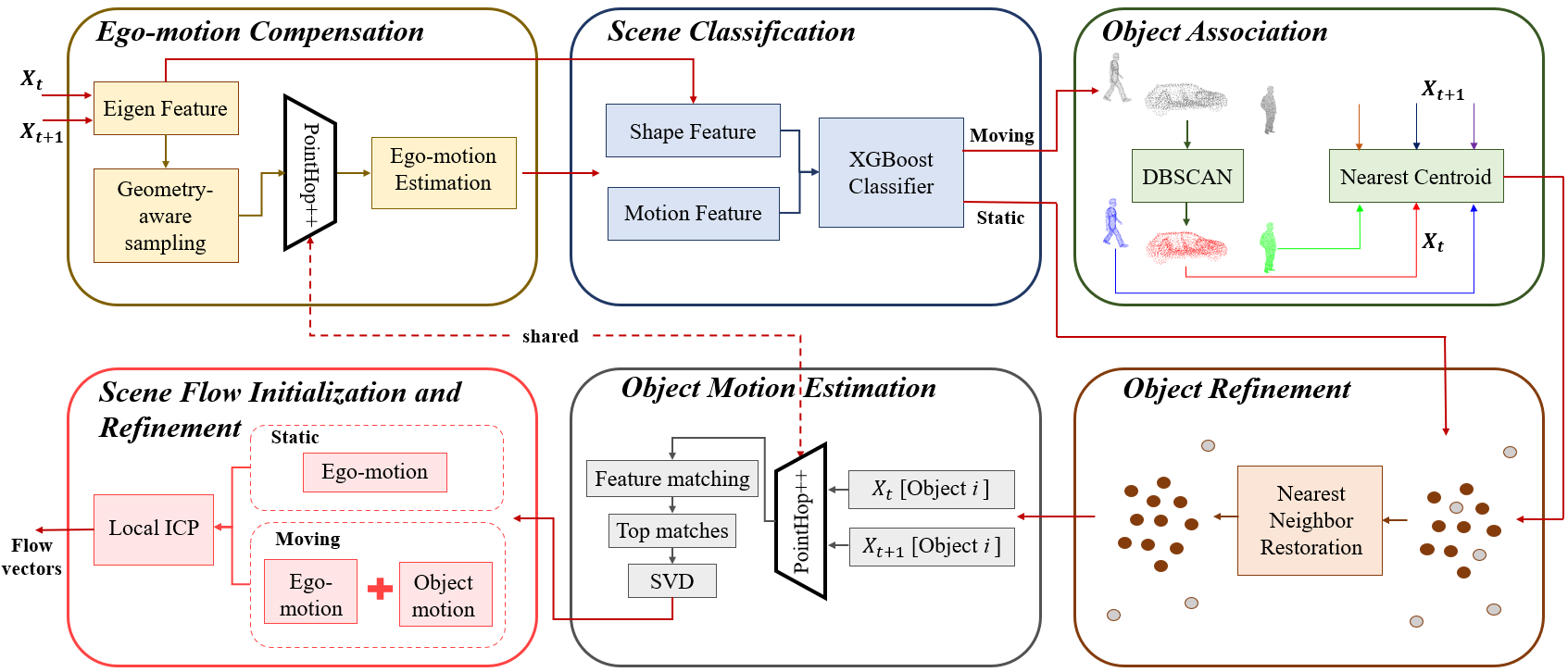}
\caption{An overview of the PointFlowHop method, which consists of six
modules: 1) ego-motion compensation, 2) scene classification, 3) object
association, 4) object refinement, 5) object motion estimation, and 6)
scene flow initialization and refinement.} \label{fig:architecture}
\end{figure*}
%%%%%%%%%%%%%%%%%%%%%%%%%%%%%%%%%%%%%%%%%%%%%%%%%%%%%%%%%%%%%%%%%%%%%

\section{Proposed PointFlowHop Method}\label{sec:method}

The system diagram of the proposed PointFlowHop method is shown in Fig.
\ref{fig:architecture}. It takes two consecutive point clouds $X_t \in
\mathbb{R}^{n_t \times 3}$ and $X_t \in \mathbb{R}^{n_{t+1} \times 3}$
as the input and calculates the point-wise flow $\bar{f_t} \in 
\mathbb{R}^{n_1 \times 3}$ for the points in $X_t$.

PointFlowHop decomposes the scene flow estimation problem into two
subproblems: 1) determining vehicle's ego-motion $(T_{ego})$ and 2)
estimating the motion of each individual object (denoted by
$(T_{obj_i})$ for object $i$). It first proceeds by determining and
compensating the ego-motion and classifying scene points as being moving
or static in modules 1 and 2, respectively.  Next, moving points are
clustered and associated as moving objects in modules 3 and 4, and the
motion of each object is estimated in module 5. Finally, the flow
vectors of static and moving points are jointly refined.  These steps
are detailed below. 

\subsection{Module 1: Ego-motion Compensation}

The $i^{th}$ point in $X_t$ has coordinates $(x_t^i,y_t^i,z_t^i)$.
Suppose this point is observed at $(x_{t+1}^i,y_{t+1}^i,z_{t+1}^i)$ in
$X_{t+1}$. These point coordinates are expressed in the respective LiDAR
coordinate systems centered at the vehicle position at time $t$ and
$t+1$. Since the two coordinate systems may not overlap due to vehicle's
motion, the scene flow vector, $\bar{f_t}^i$, of the $i^{th}$ point
cannot be simply calculated using vector difference.  Hence, we begin by
aligning the two coordinates systems or, in other words, we compensate
for the vehicle motion (or called ego-motion). 

\begin{figure*}[!t]
\centering
\includegraphics[scale=0.65]{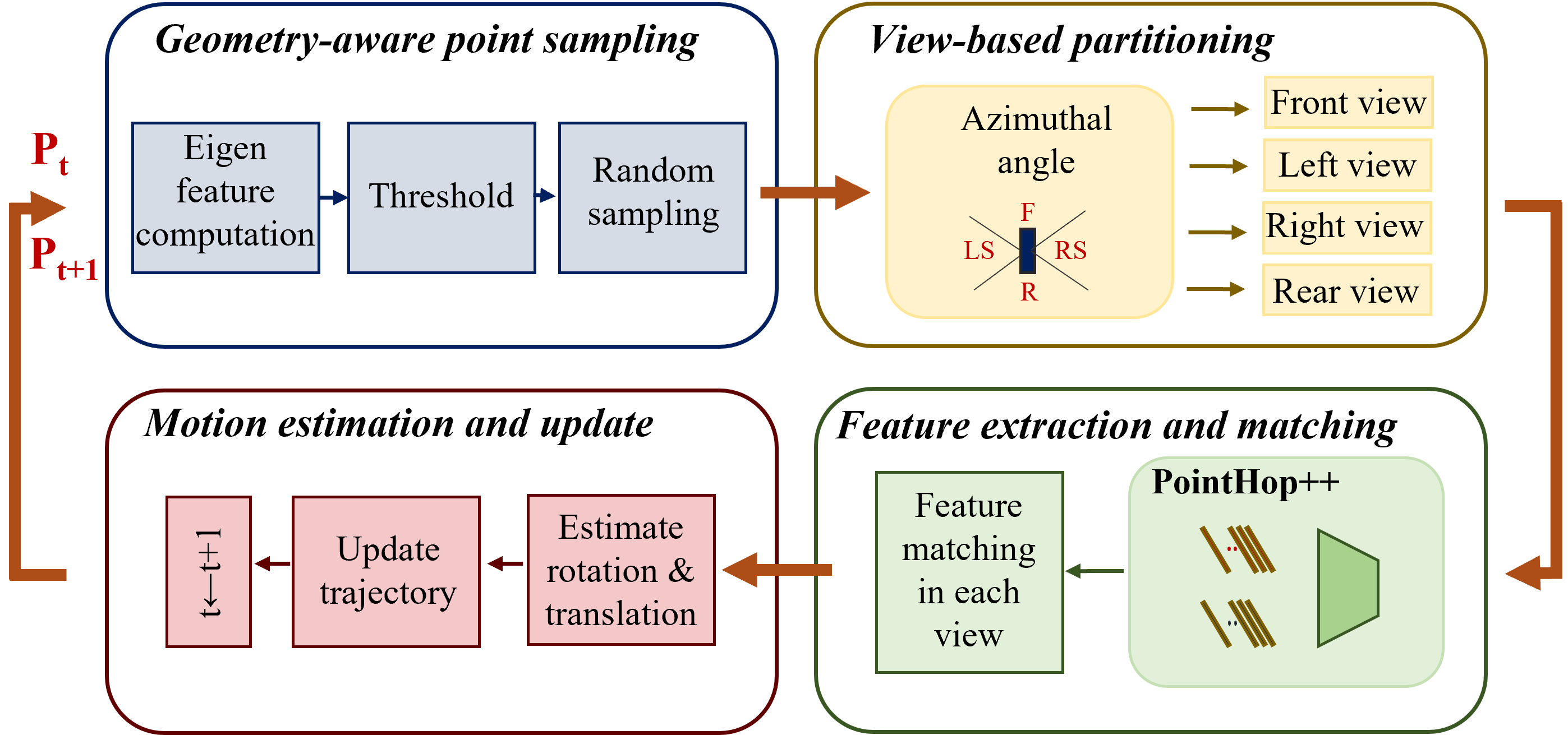}
\caption{An overview of the GreenPCO method \cite{kadam2022greenpco}.} \label{fig:greenpco}
\end{figure*}

The ego-motion compensation module in PointFlowHop is built upon a
recently proposed point cloud odometry estimation method, called
GreenPCO \cite{kadam2022greenpco}. It is briefly reviewed below for
self-containedness. GreenPCO determines the vehicle trajectory
incrementally by analyzing consecutive point cloud scans. It is
conducted with the following four steps. First, the two point clouds are
sampled using the geometry-aware sampling method, which selects points
spatially spread out with salient local surfaces based on the eigen
features \cite{hackel2016fast}. Second, the sampled points from the two
point clouds are divided into four views - front, left, right and rear
based on the azimuthal angles. Third, point features are extracted using
PointHop++ \cite{zhang2020pointhop++}. The features are used to find
matching points between the two point clouds in each view.  Last, the
pairs of matched points are used to estimate the vehicle trajectory.
These steps are repeated as the vehicle advances in the environment. The
diagram of the GreenPCO method is depicted in Fig.  \ref{fig:greenpco}.

Ego-motion estimation in PointFlowHop involves a
single iteration of GreenPCO whereby the vehicle's motion from time $t$
to $t+1$ is estimated. Then, the ego-motion can be represented by the 3D
transformation, $T_{ego}$, which consists of a 3D rotation and 3D
translation. Afterward, we use $T_{ego}$ to warp $X_t$ to $\Tilde{X}_{t}$, making it
in the same coordinate system as that of $X_{t+1}$. Then, the flow vector 
can be computed by
\begin{equation}\label{eq:flow_vector}
\bar{f_t}^i=(x_{t+1}^i-\Tilde{x}_t^i,y_{t+1}^i-\Tilde{y}_t^i,z_{t+1}^i 
-\Tilde{z}_t^i), 
\end{equation}
where $(\Tilde{x}_t^i,\Tilde{y}_t^i,\Tilde{z}_t^i)$ is the warped
coordinate of the $i^{th}$ point. 

%%%%%%%%%%%%%%%%%%%%%%%%%%%%%%%%%%%%%%%%%%%%%%%%%%%%%%%%%%%%%%%%%%%%%
%%%%%%%%%%%%%%%%%%%%%%%%%%%%%%%%%%%%%%%%%%%%%%%%%%%%%%%%%%%%%%%%%%%%%

\subsection{Module 2: Scene Classification}

After compensating for ego-motion, the resulting $\Tilde{X}_t$ and
$X_{t+1}$ are in the same coordinate system (i.e., that of $X_{t+1}$).
Next, we coarsely classify scene points in $\Tilde{X}_t$ and $X_{t+1}$
into moving and static two classes. Generally speaking, the moving
points may belong to objects such as cars, pedestrians, mopeds, etc.,
while the static points correspond to objects like buildings, poles,
etc. The scene flow of moving points can be analyzed later while static
points can be assigned a zero flow (or equal to the ego-motion depending
on the convention of the coordinate systems used).  This means that the
later stages of PointFlowHop would process fewer points. 

For the scene classifier, we define a set of shape and motion features
that are useful in distinguishing static and moving points. These features
are explained below.
\begin{itemize}
\item Shape features \\
We reuse the eigen features \cite{hackel2016fast} calculated in the
ego-motion estimation step. They summarize the distribution of
neighborhood points using covariance analysis. The analysis provides a
4-dimensional feature vector comprising of linearity, planarity, eigen
sum and eigen entropy. 
\item Motion feature \\
We first voxelize $\Tilde{X}_t$ and $X_{t+1}$ with a voxel size of 2 meters.
Then, the motion feature for each point in $\Tilde{X}_t$ is the distance
to the nearest voxel center in $X_{t+1}$, and vice versa, for each point
in $X_{t+1}$. 
\end{itemize}

The 5-dimensional (shape and motion) feature vector is fed to a binary
XGBoost classifier. For training, we use the point-wise class labels
provided by the SemanticKITTI \cite{behley2019iccv} dataset.  We observe
that the 5D shape/motion feature vector are sufficient for decent
classification. The classification accuracy on the SemanticKITTI dataset
is 98.82\%.  Furthermore, some of misclassified moving points are
reclassified in the subsequent object refinement step.

\subsection{Module 3: Object Association} 

\begin{figure}[!t]
\centering
\includegraphics[scale=0.525]{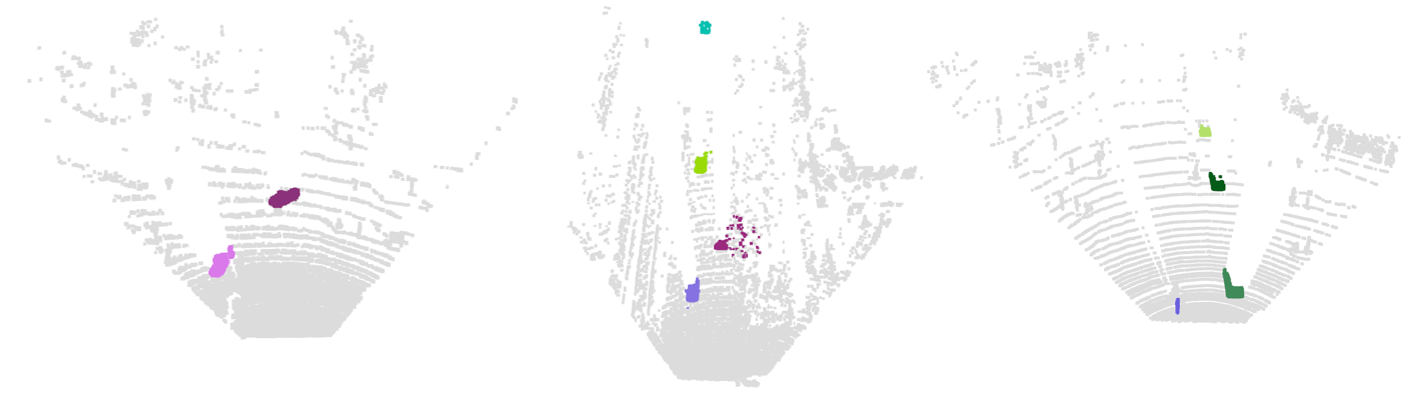}
\caption{Objects clustered using the DBSCAN algorithm are shown
in different colors.}\label{fig:clusters} 
\end{figure}

We simplify the problem of motion analysis on moving points by grouping
moving points into moving objects. To discover objects from moving
points, we use the Density-based Spatial Clustering for Applications
with Noise (DBSCAN) \cite{ester1996density} algorithm. Simply speaking,
DBSCAN iteratively clusters points based on the minimum distance ($eps$)
and the minimum points ($minPts$) parameters. Parameter $eps$ gives the
minimum Euclidean distance between points considered as neighbors.
Parameter $minPts$ determines the minimum number of points to form a
cluster.  Some examples of the objects discovered using PointFlowHop are
colored in Fig. \ref{fig:clusters}. 

Points belonging to distinct objects may get clustered together. We put
the points marked as ``outliers'' by DBSCAN in the set of static points.
The DBSCAN algorithm is run on $\Tilde{X}_t$ and $X_{t+1}$ separately.
Later, we use cluster centroids to associate objects between
$\Tilde{X}_t$ and $X_{t+1}$. That is, for each centroid in
$\Tilde{X}_t$, we locate its nearest centroid in $X_{t+1}$. 

\subsection{Module 4: Object Refinement} 

Next, we perform an additional refinement step to recover some of the
misclassified points during shape classification and potential inlier
points during object association. This is done using the nearest
neighbor rule within a defined radius neighborhood. For each point
classified as a moving point, we re-classify static points lying within
the neighborhood as moving points. The object refinement operation is
conducted on $\Tilde{X}_t$ and $X_{t+1}$. 

The refinement step is essential for two reasons. First, an imbalance
class distribution between static and moving points usually leads to the
XGBoost classifier to favor the dominant class (which is the static
points). Then, the precision and recall for moving points are still low
in spite of high classification accuracy. Second, in the clustering
step, it is difficult to select good values for $eps$ and $minPts$ that
are robust in all scenarios for the sparse LiDAR point clouds.  This may
lead to some points being marked as outliers by DBSCAN.  Overall, the
performance gains of our method reported in Sec.  \ref{sec:experiments}
are a result of the combination of all steps and not due to a single
step in particular. 

%%%%%%%%%%%%%%%%%%%%%%%%%%%%%%%%%%%%%%%%%%%%%%%%%%%%%%%%%%%%%%%%%%%%%
%%%%%%%%%%%%%%%%%%%%%%%%%%%%%%%%%%%%%%%%%%%%%%%%%%%%%%%%%%%%%%%%%%%%%

\subsection{Module 5: Object Motion Estimation}

\begin{figure*}[!t]
\centering
\includegraphics[scale=0.45]{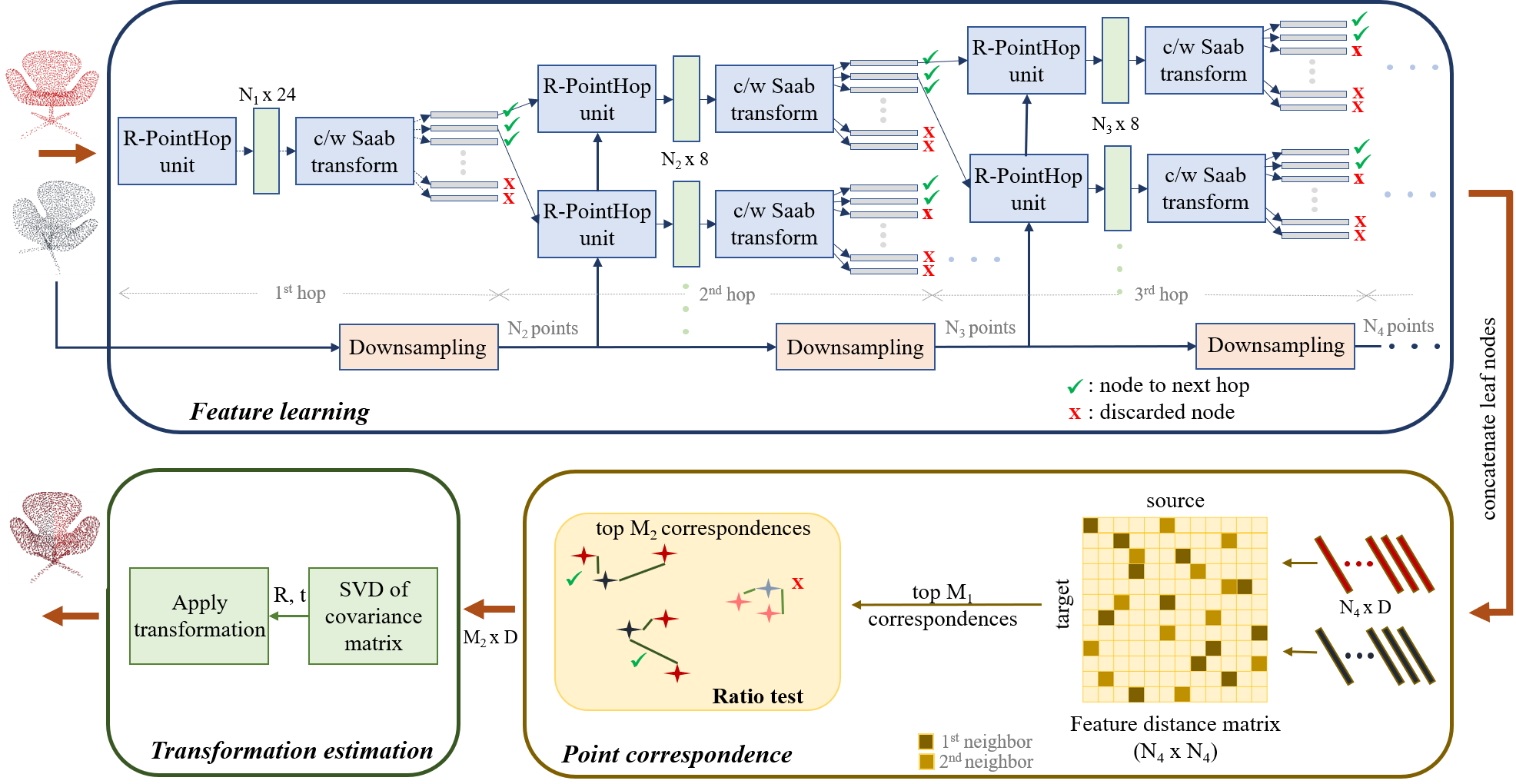}
\caption{An overview of the R-PointHop method \cite{kadam2022r}.} \label{fig:rpointhop}
\end{figure*}

We determine the motion between each pair of associated objects in this
step. For that, we follow a similar approach as taken by a point cloud
rigid registration method, R-PointHop \cite{kadam2022r}.  The objective
of R-PointHop is to register the source point cloud with the target
point cloud.  The block diagram of R-PointHop is illustrated in Fig.
\ref{fig:rpointhop}. It includes the following two major steps.  First,
the source and target point clouds are fed to a sequence of R-PointHop
units for hierarchical feature learning (or multiple hops) in the
feature learning step. Point clouds are downsampled between two hops by
iteratively selecting farther points. The R-PointHop unit comprises of
constructing a local point descriptor followed by the channel-wise Saab
transform \cite{chen2020pixelhop++}.  Second, the point features are
used to find pairs of corresponding points. The optimal rigid
transformation that aligns the two point clouds is then solved as a
energy minimization problem \cite{schonemann1966generalized}. 

For object motion estimation in PointFlowHop, the features of refined
moving points from $\Tilde{X}_t$ and $X_{t+1}$ are extracted using the
trained PointHop++ model. We reuse the same model from the ego-motion
estimation step here. While four hops with intermediate downsampling is
used in R-PointHop, the PointHop++ model in PointFlowHop only involves
two hops without downsampling to suit the LiDAR data.  Since
$\Tilde{X}_{t}^{obj_i}$ and $X_{t+1}^{obj_i}$ are two sets of points
belonging to object $i$, we find corresponding points between the two
point clouds using the nearest neighbor search in the feature space.
The correspondence set is further refined by selecting top
correspondences based on: 1) the minimum feature distance criterion and
2) the ratio test (the minimum ratio of the distance between the first
and second best corresponding points).  The refined correspondence set
is then used to estimate the object motion as follows.

First, the mean coordinates of the corresponding points in
$\Tilde{X}_{t}^{obj_i}$ and $X_{t+1}^{obj_i}$ are found by:
\begin{equation}
\Bar x_t^{obj_i}=\frac{1}{N_{obj_i}}\sum\limits_{j=1}^{N_{obj_i}}
\Tilde{x}_t^{obj_{ij}}, \quad \Bar
x_{t+1}^{obj_i}=\frac{1}{N_{obj_i}}\sum\limits_{j=1}^{N_{obj_i}}
x_{t+1}^{obj_{ij}}.
\end{equation}
Then, the $3\times 3$ covariance matrix is computed using the pairs of
corresponding points as
\begin{equation}
K(\Tilde{X}_{t}^{obj_i},X_{t+1}^{obj_i})=\sum\limits_{j=1}^{N_{obj_i}}
(\Tilde{x}_t^{obj_{ij}}-\Bar x_t^{obj_i})(x_{t+1}^{obj_{ij}}-\Bar x_{t+1}^{obj_i})^T. 
\end{equation}
The Singular Value Decomposition of $K$ gives matrices $U$ and $U$, which 
are formed by the left and right singular vectors, respectively. Mathematically,
we have
\begin{equation}
K(\Tilde{X}_{t}^{obj_i},X_{t+1}^{obj_i})= USV^T.
\end{equation}
Following the orthogonal procrustes formulation \cite{schonemann1966generalized}, 
the optimal motion of $\Tilde{X}_{t}^{obj_i}$ can be expressed in form of a
rotation matrix $R^{obj_i}$ and a translational vector $t^{obj_i}$. They
can be computed as
\begin{equation}
R^{obj_i}=VU^T, \quad t^{obj_i}=-R^{obj_i}\Bar x_t^{obj_i}+\Bar x_{t+1}^{obj_i}.
\end{equation}
Since $(R^{obj_i}, t^{obj_i})$ form the object motion model for object $i$,
it is denoted as $T_{obj_i}$. 

Actually, once we find the corresponding point $x_{t+1}^{obj_{ij}}$ 
of $\Tilde{x}_t^{obj_{ij}}$, the flow vector may be set to
$$
\Tilde{x}_t^{obj_{ij}} = x_{t+1}^{obj_{ij}}-\Tilde{x}_t^{obj_{ij}}.
$$ 
However, this point-wise flow vector can be too noisy, and it is desired
to use a flow model for the object rather than each point.  The object
flow model found using SVD in the step after finding correspondences is
optimal in the mean square sense over all corresponding points and,
hence, is more robust. It makes a reasonable assumption of existence of
a rigid transformation between the two objects. 

%%%%%%%%%%%%%%%%%%%%%%%%%%%%%%%%%%%%%%%%%%%%%%%%%%%%%%%%%%%%%%%%%%%%%
%%%%%%%%%%%%%%%%%%%%%%%%%%%%%%%%%%%%%%%%%%%%%%%%%%%%%%%%%%%%%%%%%%%%%

\subsection{Module 6: Flow Initialization and Refinement}

In the last module, we apply the object motion model $T_{obj_i}$ to
$\Tilde{X}_{t}^{obj_i}$ and align it with ${X}_{t+1}^{obj_i}$. Since the
static points do not have any motion, they are not further transformed.
We denote the new transformed point cloud as $\Tilde{X}_{t}'$. At this
point, we have obtained an initial estimate of the scene flow for each
point in $X_t$.  For static points, the flow is given by the ego-motion
transformation $T_{ego}$. For the moving points, it is a composition of
ego motion and corresponding object's motion $T_{ego}\cdot T_{obj_i}$. 

%%%%%%%%%%%%%%%%%%%%%%%%%%%%%%%%%%%%%%%%%%%%%%%%%%%%%%%%%%%%%%%%%%%%%
\begin{figure}[!t]
\centering
\includegraphics[scale=0.72]{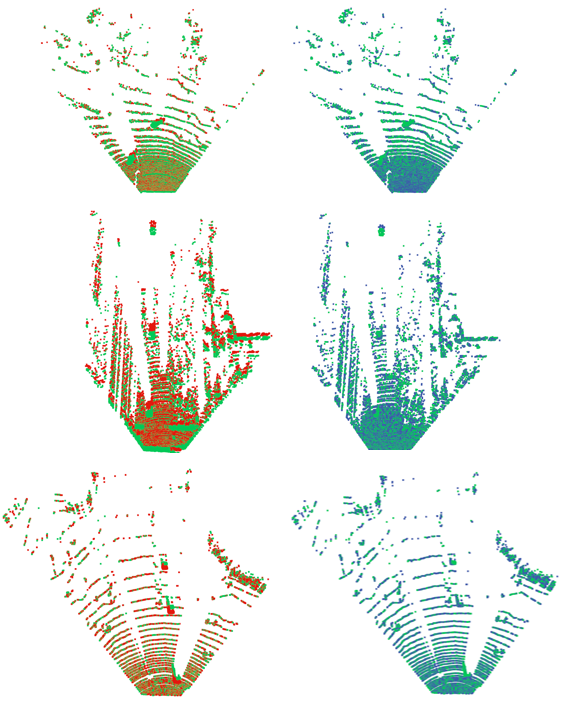}
\caption{Flow estimation results using PointFlowHop: input point
clouds (left) and warped output using flow vectors (right).} \label{fig:flow}
\end{figure}
%%%%%%%%%%%%%%%%%%%%%%%%%%%%%%%%%%%%%%%%%%%%%%%%%%%%%%%%%%%%%%%%%%%%%

In this module, we refine the flow for all points in $\Tilde{X}_{t}'$
using the Iterative Closest Point (ICP) \cite{besl1992method} algorithm
in small non-overlapping regions. In each region, the points in
$\Tilde{X}_{t}'$ falling within it are aligned with corresponding points
in ${X}_{t+1}$. The flow refinement step ensures a tighter alignment and
is a common post processing operation in several related tasks. Finally,
the flow vectors for ${X}_{t}$ are calculated as difference between the
transformed and initial coordinates. Exemplar pairs of input and scene
flow compensated point clouds using PointFlowHop are shown in Fig.
\ref{fig:flow}. 

\section{Experiments}\label{sec:experiments}

In this section, we report experimental results on real world LiDAR
point cloud datasets. We choose the stereoKITTI \cite{menze2015joint,
menze2018object} and the Argoverse \cite{chang2019argoverse} two datasets
since they represent challenging scenes in autonomous driving
environments.  StereoKITTI has 142 pairs of point clouds. The ground
truth flow of each pair is derived from the 2D disparity maps and the
optical flow information. There are 212 test samples for Argoverse whose
flow annotations were given in \cite{pontes2020scene}. We use per-point
labels from the SemanticKITTI dataset \cite{behley2019iccv} to train our
scene classifier. 

Following series of prior art, we measure the performance in the following
metrics:
\begin{itemize}
\item {\em 3D end point error (EPE3D).} It is the mean Euclidean distance
between the estimated and the ground truth flow. 
\item {\em Strict accuracy (Acc3DS).} It is the percentage of points for
which EPE3D is less than 0.05m or the relative error is less than 0.05.
\item {\em Relaxed accuracy (Acc3DR).} It gives the ratio of points for
which EPE3D is less than 0.1m or the relative error is less than 0.1. 
\item {\em Percentage of Outliers.} It is the ratio of points
for which EPE3D is greater than 0.3m or the relative error is greater
than 0.1. This is reported for the StereoKITTI dataset only. 
\item {\em Mean angle error (MAE).} It is the mean of the angle errors
between the estimated and the ground truth flow of all points 
expressed in the unit of radians. This is reported for the Argoverse 
dataset only. 
\end{itemize}

\subsection{Performance Benchmarking}

The scene flow estimation results on stereoKITTI and Argoverse are
reported in Table \ref{tab:kitti} and Table \ref{tab:argoverse},
respectively. For comparison, we show the performance of several
representative methods proposed in the past few years.  Overall, the
EPE3D, Acc3DS and Acc3DR values are significantly better for stereoKITTI
as compared to the Argoverse dataset. This is because Argoverse is a
more challenging dataset.  Furthermore, PointFlowHop outperforms all
benchmarking methods in almost all evaluation metrics on both datasets. 

%%%%%%%%%%%%%%%%%%%%%%%%%%%%%%%%%%%%%%%%%%%%%%%%%%%%%%%%%%%%%%%%%%%%
\begin{table}[htbp]
\centering
\caption{Comparison of scene flow estimation results on the Stereo 
KITTI dataset, where the best performance number is shown in 
boldface.} \label{tab:kitti}
\renewcommand\arraystretch{1.3}
\newcommand{\tabincell}[2]{\begin{tabular}{@{}#1@{}}#2\end{tabular}}
% \resizebox{\columnwidth}{!}{
\begin{tabular}{c | c c  c c  c c  c c} \hline 
Method & EPE3D (m)$\downarrow$ & Acc3DS $\uparrow$ & Acc3DR $\uparrow$ & Outliers $\downarrow$ \\ \hline
FlowNet3D \cite{liu2019flownet3d}  & 0.177 & 0.374  & 0.668 & 0.527  \\ 
HPLFlowNet \cite{gu2019hplflownet} & 0.117 & 0.478  & 0.778 & 0.410  \\ 
PointPWC-Net \cite{wu2020pointpwc}  & 0.069 & 0.728  & 0.888 & 0.265 \\
FLOT \cite{puy2020flot}  & 0.056 & 0.755 & 0.908 & 0.242             \\ 
HALFlow \cite{wang2021hierarchical}  & 0.062 & 0.765  & 0.903 & 0.249 \\
Rigid3DSceneFlow \cite{gojcic2021weakly} & 0.042 & 0.849  & 0.959 & 0.208 \\
PointFlowHop (Ours) & \bf{0.037} & \bf{0.938}  & \bf{0.974} & \bf{0.189}  \\ \hline
\end{tabular}
% }
\end{table}
%%%%%%%%%%%%%%%%%%%%%%%%%%%%%%%%%%%%%%%%%%%%%%%%%%%%%%%%%%%%%%%%%%%%

%%%%%%%%%%%%%%%%%%%%%%%%%%%%%%%%%%%%%%%%%%%%%%%%%%%%%%%%%%%%%%%%%%%%
\begin{table}[htbp]
\centering
\caption{Comparison of scene flow estimation results on the Argoverse 
dataset, where the best performance number is shown in boldface.} 
\label{tab:argoverse}
\renewcommand\arraystretch{1.3}
\newcommand{\tabincell}[2]{\begin{tabular}{@{}#1@{}}#2\end{tabular}}
% \resizebox{\columnwidth}{!}{
\begin{tabular}{c | c c  c c  c c  c c} \hline 
Method & EPE3D (m) $\downarrow$ & Acc3DS $\uparrow$ & Acc3DR $\uparrow$ & MAE (rad) $\downarrow$ \\ \hline
FlowNet3D \cite{liu2019flownet3d}  & 0.455 & 0.01  & 0.06 & 0.736          \\ 
PointPWC-Net \cite{wu2020pointpwc}  & 0.405 & 0.08  & 0.25 & 0.674         \\
Just Go with the Flow \cite{mittal2020just} & 0.542 & 0.08 & 0.20 & 0.715  \\
NICP \cite{amberg2007optimal} & 0.461 & 0.04  & 0.14 & 0.741               \\ 
Graph Laplacian \cite{pontes2020scene} & 0.257 & 0.25  & 0.48 & 0.467      \\ 
Neural Prior \cite{li2021neural} & 0.159 & 0.38  & 0.63 & \bf{0.374}       \\
PointFlowHop (Ours) & \bf{0.134} & \bf{0.39}  & \bf{0.71} & 0.398          \\ \hline
\end{tabular}
% }
\end{table}
%%%%%%%%%%%%%%%%%%%%%%%%%%%%%%%%%%%%%%%%%%%%%%%%%%%%%%%%%%%%%%%%%%%%

\subsection{Ablation Study} 

In this section, we assess the role played by each individual module
of PointFlowHop using the stereo KITTI dataset as an example. 

{\bf Ego-motion compensation.} First, we may replace GreenPCO
\cite{kadam2022greenpco} with ICP \cite{besl1992method} for ego-motion
compensation. The results are presented in Table
\ref{tab:ablation_study_ego}. We see a sharp decline in performance with
ICP. The substitution makes the new method much worse than all
benchmarking methods. While the naive ICP could be replaced with other
advanced model-free methods, it is preferred to use GreenPCO since the
trained PointHop++ model is still needed later. 

%%%%%%%%%%%%%%%%%%%%%%%%%%%%%%%%%%%%%%%%%%%%%%%%%%%%%%%%%%%%%%%%%%%%
\begin{table}[htbp]
\centering
\caption{Ego-motion compensation -- ICP vs. GreenPCO.} \label{tab:ablation_study_ego}
\renewcommand\arraystretch{1.3}
\newcommand{\tabincell}[2]{\begin{tabular}{@{}#1@{}}#2\end{tabular}}
% \resizebox{\columnwidth}{!}{
\begin{tabular}{c | c c  c c  c c  c c} \hline 
Ego-motion Method & EPE3D $\downarrow$ & Acc3DS $\uparrow$ & Acc3DR $\uparrow$ & Outliers $\downarrow$ \\ \hline
ICP \cite{besl1992method} & 0.574 & 0.415  & 0.481 & 0.684  \\ \hline
GreenPCO \cite{kadam2022greenpco} & \bf{0.037} & \bf{0.938}  & \bf{0.974} & \bf{0.189}  \\  \hline
\end{tabular}
% }
\end{table}
%%%%%%%%%%%%%%%%%%%%%%%%%%%%%%%%%%%%%%%%%%%%%%%%%%%%%%%%%%%%%%%%%%%%

%%%%%%%%%%%%%%%%%%%%%%%%%%%%%%%%%%%%%%%%%%%%%%%%%%%%%%%%%%%%%%%%%%%%
\begin{table}[htbp]
\centering
\caption{Performance gain due to object refinement.} \label{tab:ablation_study_object}
\renewcommand\arraystretch{1.3}
\newcommand{\tabincell}[2]{\begin{tabular}{@{}#1@{}}#2\end{tabular}}
% \resizebox{\columnwidth}{!}{
\begin{tabular}{c | c c  c c  c c  c c} \hline 
Object Refinement & EPE3D $\downarrow$ & Acc3DS $\uparrow$ & Acc3DR $\uparrow$ & Outliers $\downarrow$ \\ \hline
\xmark  & 0.062 & 0.918 & 0.947 & 0.208  \\ \hline
\cmark & \bf{0.037} & \bf{0.938}  & \bf{0.974} & \bf{0.189}  \\  \hline
\end{tabular}
% }
\end{table}
%%%%%%%%%%%%%%%%%%%%%%%%%%%%%%%%%%%%%%%%%%%%%%%%%%%%%%%%%%%%%%%%%%%%

%%%%%%%%%%%%%%%%%%%%%%%%%%%%%%%%%%%%%%%%%%%%%%%%%%%%%%%%%%%%%%%%%%%%
\begin{table}[htbp]
\centering
\caption{Performance gain due to flow refinement.} \label{tab:ablation_study_flow}
\renewcommand\arraystretch{1.3}
\newcommand{\tabincell}[2]{\begin{tabular}{@{}#1@{}}#2\end{tabular}}
% \resizebox{\columnwidth}{!}{
\begin{tabular}{c | c c  c c  c c  c c} \hline 
Flow Refinement & EPE3D $\downarrow$ & Acc3DS $\uparrow$ & Acc3DR $\uparrow$ & Outliers $\downarrow$ \\ \hline
\xmark  & 0.054 & 0.862  & 0.936 & 0.230  \\ \hline
\cmark  & \bf{0.037} & \bf{0.938}  & \bf{0.974} & \bf{0.189}  \\  \hline
\end{tabular}
% }
\end{table}
%%%%%%%%%%%%%%%%%%%%%%%%%%%%%%%%%%%%%%%%%%%%%%%%%%%%%%%%%%%%%%%%%%%%

{\bf Performance Gain Due to Object Refinement.} Next, we compare
PointFlowHop with and without the object refinement step. The results
are shown in Table \ref{tab:ablation_study_object}. We see consistent
performance improvement in all evaluation metrics with the object
refinement step.  On the other hand, the performance of PointFlowHop is
still better than that of benchmarking methods except for
Rigid3DSceneFLow \cite{gojcic2021weakly} (see Table \ref{tab:kitti})
even without object refinement. 

{\bf Performance Gain Due to Flow Refinement.} Finally, we compare
PointFlowHop with and without the flow refinement step in Table
\ref{tab:ablation_study_flow}. It is not surprising that flow refinement
is crucial in PointFlowHop. However, one may argue the refinement step
may be included in any of the discussed methods as a post processing
operation. While this argument is valid, we see that even without flow
refinement, PointFlowHop still is better than almost all methods (see
Table \ref{tab:kitti}). Between object refinement and flow refinement,
flow refinement seems slightly more important if we consider all four
evaluation metrics jointly.

\subsection{Complexity Analysis}

The complexity of a machine learning method can be examined from
multiple angles, including training time, the number of model parameters
(i.e., the model size) and the number of floating point operations
(FLOPs) during inference. These metrics are valuable besides performance
measures such as prediction accuracy/error. Furthermore, since some
model-free methods (e.g., LOAM \cite{zhang2014loam}) and the recently
proposed KISS-ICP \cite{vizzo2023kiss} can offer state-of-the-art
results for related tasks such as Odometry and Simultaneous Localization
and Mapping (SLAM), the complexity of learning-based methods deserves
additional attention.

To this end, PointFlowHop offers impressive benefits as compared to
representative DL-based solutions. Training in PointFlowHop only
involves the ego-motion compensation and shape classification steps.
For object motion estimation, PointHop++ obtained from the ego-motion
compensation step is reused while the rest of the operations in
PointFlowHop are parameter-free and performed only in inference. 

Table \ref{tab:parameters_time} provides details about the number of
parameters of PointFlowHop.  It adopts the PointHop++ architecture with
two hops. The first hop has 13 kernels of dimension 88 while the second
hop has 104 kernels of dimension 8. For XGBoost, it has 100 decision
tree estimators, each of which has a maximum depth of 3.  We also report
the training time of PointFlowHop in the same table, where the training
is conducted on Intel(R) Xeon(R) CPU E5-2620 v3 at 2.40GHz. 

%%%%%%%%%%%%%%%%%%%%%%%%%%%%%%%%%%%%%%%%%%%%%%%%%%%%%%%%%%%%%%%%%%%%
\begin{table}[htbp]
\centering
\caption{The number of trainable parameters and training time of the 
proposed PointFlowHop.} \label{tab:parameters_time}
\renewcommand\arraystretch{1.3}
\newcommand{\tabincell}[2]{\begin{tabular}{@{}#1@{}}#2\end{tabular}}
% \resizebox{\columnwidth}{!}{
\begin{tabular}{c | c | c} \hline 
 & Number of Parameters & Training time \\ \hline
Hop 1  & 1144 &  \multirow{2}{*}{20 minutes} \\ 
Hop 2  & 832 &   \\ \hline
XGBoost  & 2200 & 12 minutes   \\ \hline
\bf{Total} & \bf{4176} & \bf{32 minutes}   \\ \hline
\end{tabular}
% }
\end{table}
%%%%%%%%%%%%%%%%%%%%%%%%%%%%%%%%%%%%%%%%%%%%%%%%%%%%%%%%%%%%%%%%%%%

While we do not measure the training time of other methods ourselves, we
use \cite{pontes2020scene} as a reference to compare our training time
with others. It took the authors of \cite{pontes2020scene} about 18
hours to train and fine-tune the FlowNet3D \cite{liu2019flownet3d}
method for the KITTI dataset and about 3 days for the Argoverse dataset.
We expect comparable time for other methods. Thus, PointFlowHop is
extremely efficient in this context. While the Graph Laplacian
method \cite{pontes2020scene} offers a variant where the scene flow is
entirely optimized at runtime (non-learning based), its performance
is inferior to ours as shown in Table \ref{tab:argoverse}. 

%%%%%%%%%%%%%%%%%%%%%%%%%%%%%%%%%%%%%%%%%%%%%%%%%%%%%%%%%%%%%%%%%%%%
\begin{table}[htbp]
\centering
\caption{Comparison of model sizes (in terms of the number of parameters) 
and computational complexity of inference (in terms of FLOPs) of four 
benchmarking methods.}\label{tab:complexity}
\renewcommand\arraystretch{1.3}
\newcommand{\tabincell}[2]{\begin{tabular}{@{}#1@{}}#2\end{tabular}}
% \resizebox{\columnwidth}{!}{
\begin{tabular}{c | c | c} \hline 
Method & Number of Parameters & FLOPs \\ \hline
FlowNet3D \cite{liu2019flownet3d}  & 1.23 M (308X)  &  11.67 G (61X)   \\ 
PointPWC Net \cite{wu2020pointpwc} & 7.72 M (1930X) &  17.46 G (92X)   \\
FLOT \cite{puy2020flot}            & 110 K (28X)    &  54.65 G (288X)  \\ 
PointFlowHop (Ours)                & \bf{4 K} (1X)  &  \bf{190 M} (1X) \\  \hline
\end{tabular}
% }
\end{table}
%%%%%%%%%%%%%%%%%%%%%%%%%%%%%%%%%%%%%%%%%%%%%%%%%%%%%%%%%%%%%%%%%%%

Finally, we compare the model sizes and computational complexity of four
benchmarking methods in Table \ref{tab:complexity}.  It is apparent that
PointFlowHop demands significantly less parameters than other methods.
Furthermore, we compute the number of floating-point operations (FLOPs)
of PointFlowHop analytically during inference and report it in Table
\ref{tab:complexity}. While calculating the FLOPs, we consider input
point clouds containing 8,192 points. Thus, the normalized FLOPs per
point is 23.19K.  We conclude from the above discussion that
PointFlowHop offers a green and high-performance solution to 3D scene
flow estimation.

\section{Conclusion and Future Work}\label{sec:conclusion}

A green and interpretable 3D scene flow estimation method called
PointFlowHop was proposed in this work. PointFlowHop takes two
consecutive LiDAR point cloud scans and determines the flow vectors for
all points in the first scan. It decomposes the flow into vehicle's
ego-motion and the motion of an individual object in the scene.  The
superior performance of PointFlowHop over benchmarking DL-based methods
was demonstrated on stereoKITTI and Argoverse datasets.  Furthermore,
PointFlowHop has advantages in fewer trainable parameters and fewer
FLOPs during inference. 

One future research direction is to extend PointFlowHop for the 3D
object detection task. Along this line, we may detect moving objects
using PointFlowHop and derive 3D bounding boxes around them. The
clustered points obtained by PointFlowHop may act as an initialization
in the object detection process. Another interesting problem to pursue
is simultaneous flow estimation and semantic segmentation. The
task-agnostic nature of our representation learning can be useful. 

\section*{Acknowledgments} 

This work was supported by a research gift from Tencent Media Lab. The
authors also acknowledge the Center for Advanced Research Computing
(CARC) at the University of Southern California for providing computing
resources that have contributed to the research results reported within
this publication. URL: https://carc.usc.edu. 

\bibliographystyle{unsrt}
\bibliography{refs}

\end{document}